\documentclass{article}

\usepackage{PRIMEarxiv}

\usepackage[utf8]{inputenc} 
\usepackage[T1]{fontenc}    
\usepackage{hyperref}       
\usepackage{url}            
\usepackage{booktabs}       
\usepackage{amsfonts}       
\usepackage{nicefrac}       
\usepackage{microtype}      
\usepackage{lipsum}
\usepackage{fancyhdr}       
\usepackage{graphicx}       
\graphicspath{{media/}}     
\usepackage{float}
\usepackage{graphicx}
\usepackage{subcaption}
\usepackage{amsmath}
\usepackage{amssymb}
\usepackage{pifont}
\usepackage{verbatim}
\usepackage{multirow}
\usepackage{xcolor}
\usepackage{nameref}
\usepackage{CJKutf8}

\title{Is ChatGPT Involved in Texts? Measure the Polish Ratio to Detect ChatGPT-Generated Text
}

\author{
  Lingyi Yang\textsuperscript{\rm 1},
    Feng Jiang \textsuperscript{\rm 1 \rm 2 \rm 3}\thanks{Corresponding Author},
    Haizhou Li \textsuperscript{\rm 1 \rm 2} \\
  \textsuperscript{\rm 1} School of Data Science, The Chinese University of Hong Kong, Shenzhen, China\\
    \textsuperscript{\rm 2} Shenzhen Research Institute of Big Data, China\\
    \textsuperscript{\rm 3} School of Information Science and Technology, University of Science and Technology of China, China\\
  \texttt{lingyiyang@link.cuhk.edu.cn, \{jeffreyjiang,haizhouli\}@cuhk.edu.cn} \\
}

\begin{document}
\maketitle
\begin{abstract}
The remarkable capabilities of large-scale language models, such as ChatGPT, in text generation have impressed readers and spurred researchers to devise detectors to mitigate potential risks, including misinformation, phishing, and academic dishonesty. Despite this, most previous studies have been predominantly geared towards creating detectors that differentiate between purely ChatGPT-generated texts and human-authored texts. This approach, however, fails to work on discerning texts generated through human-machine collaboration, such as ChatGPT-polished texts. Addressing this gap, we introduce a novel dataset termed HPPT (ChatGPT-polished academic abstracts), facilitating the construction of more robust detectors. It diverges from extant corpora by comprising pairs of human-written and ChatGPT-polished abstracts instead of purely ChatGPT-generated texts. Additionally, we propose the "Polish Ratio" method, an innovative measure of the degree of modification made by ChatGPT compared to the original human-written text. It provides a mechanism to measure the degree of ChatGPT influence in the resulting text. Our experimental results show our proposed model has better robustness on the HPPT dataset and two existing datasets (HC3 and CDB). Furthermore, the "Polish Ratio" we proposed offers a more comprehensive explanation by quantifying the degree of ChatGPT involvement.

\end{abstract}

\section{Introduction}
\label{intro}
From GPT-2 \cite{radford2019language} to GPT-3 \cite{brown2020language}, the emergence of Large Language Models (LLM) has made it possible that machines can generate a variety of high-quality texts that are quite similar to human language, making it hard to distinguish between human-generated and AI-generated texts. The release of ChatGPT \cite{chatgpt} contributed to the widespread use of AI-generated text tools while leading to concerns that the abuse of ChatGPT may bring misleading information, phishing, and academic dishonesty~\cite{ban}. Therefore, some studies~\cite{guo2023close} attempt to build detectors based on the pre-trained language model for ChatGPT-generated texts to prevent the abuse of the powerful ChatGPT. 

However, detectors that perform well in intra-domain scenarios have poor robustness due to being easily attacked by carefully designed strategies such as paraphrasing and polishing~\cite{sadasivan2023can}, which refers to the process of revising and improving the original text. Although retrieval~\cite{krishna2023paraphrasing} can help mitigate this problem, it still poses significant challenges, such as high costs and lack of explanation. Moreover, as pointed out by the Association for Computational Linguistics (ACL), it is necessary to establish a grading system for the degree to which large models participate in the paper generation process~\footnote{\url{https://2023.aclweb.org/blog/ACL-2023-policy}}. 

In order to identify ChatGPT-polished texts and provide users with more intuitive explanations, we create a novel dataset called HPPT (ChatGPT-polished academic abstracts instead of fully generated ones) for training a detector and also propose the Polish Ratio method which measures the degree of modification made by ChatGPT compared to the original human-written text. Through experiments, we have demonstrated the effectiveness of our model in accurately identifying polished texts trained on our novel dataset. Moreover, our explanation method, the Polish Ratio, has shown promising results on both our own dataset and other datasets that have not been seen before: there are significant distinct distributions in the predicted Polish Ratio of human-written, ChatGPT-polished, and ChatGPT-generated texts.   
\textbf{Our contributions are:} a) we build a Human-ChatGPT Polished Paired abstracT (HPPT) dataset, containing polished and original paired abstracts with similarity measurements as the degree of polishing. b) We propose the Polish Ratio to help explain the detection model indicating the modification degree of the text by ChatGPT. c) The experimental results show that our model performs better than other baselines on three datasets. The code and dataset are available: \url{https://github.com/Clement1290/ChatGPT-Detection-PR-HPPT}.
\section{Related Work}
\label{rw}

Along with the appearance of large language models such as ChatGPT, some detection algorithms are proposed to prevent the abuse of such powerful AI-generated text models. Recent detection methods can be roughly grouped into two categories~\cite{tang2023science}: white-box and black-box detectors. The white-box detector needs to access the distributed probability or vocabulary of the target language model, while the black-box detector only checks the output text of the target model. Table~\ref{sum} summarizes recent algorithms that detect ChatGPT-generated texts.

\begin{table}[ht]
\resizebox{\textwidth}{!}{
\begin{tabular}{ccccc}
\hline
\textbf{Detector}                                                     & \textbf{Open Source} & \textbf{Black Box} & \textbf{Detection basis} & \textbf{Explanation}  \\ \hline
\begin{tabular}[c]{@{}c@{}}DetectGPT\\ \cite{mitchell2023detectgpt}\end{tabular}                 & \checkmark                    & \ding{55}                  & Probability Score & \textit{/} \\ \hline
\begin{tabular}[c]{@{}c@{}}Watermarks\\ \cite{kirchenbauer2023watermark}\end{tabular}                & \checkmark                    & \ding{55}                  & Red \& Green Tokens          & \textit{/}  \\ \hline
\begin{tabular}[c]{@{}c@{}}Roberta-based-detector\\ \cite{guo2023close}\end{tabular}    & \checkmark                    & \checkmark                  & Output String              & \textit{/}  \\ \hline
\begin{tabular}[c]{@{}c@{}}OpenAI\\ \cite{openAI}\end{tabular}                    & \ding{55}                    & \checkmark                  & Output String                 & \textit{/} \\ \hline
\begin{tabular}[c]{@{}c@{}}GPTZero \end{tabular}                   & \ding{55}                    & \checkmark         & Output String                & \textit{/} \\ \hline
\begin{tabular}[c]{@{}c@{}}DistilBERT-based-detector\\ \cite{mitrović2023chatgpt}\end{tabular} & \ding{55}                    & \checkmark                  & Output String               & \textit{SHAP} \\ \hline
\begin{tabular}[c]{@{}c@{}}\textbf{Ours} \end{tabular} & \checkmark                     & \checkmark                  & Output String               & \textit{Polish Ratio} \\ \hline                                               
\end{tabular}
}
\caption{Summary of recent algorithms detecting ChatGPT-generated texts.}
\label{sum}
\end{table}

\cite{gehrmann-etal-2019-gltr} is a popular white-box detector, which takes advantage of the probability and entropy of the linguistic patterns, while DetectGPT~\cite{mitchell2023detectgpt} utilizes the observation that ChatGPT texts tend to lie in areas where the log probability function has negative curvature to conduct zero-shot detection. Watermark-based detection methods~\cite{kirchenbauer2023watermark,meral2009natural} need to divide the vocabulary into a randomized set of words that is softly prompted to use during sampling ("green" tokens) and the remaining words ("red" tokens), and modify the logits embedding when the model generates the results.

Training another classifier as the detector with labeled data \cite{openAI,mitrović2023chatgpt,guo2023close} is a mainstream black-box method. \cite{guo2023close} finetunes the Roberta on the HC3 (Human ChatGPT Comparison Corpus) dataset to obtain an effective detector. However, the simple ChatGPT-generated texts in the HC3 dataset make the model trained on it vulnerable to being attacked using the polishing strategy, and the robustness is not ensured. Moreover, some effective detectors are not open-source and are directly used for commercial operations, such as GPTZero \footnote{\url{https://gptzero.me/}} and OriginalityAI~\footnote{\url{https://originality.ai/}}. 

On the other hand, the existing black-box detectors rarely provide explanations for the prediction. \cite{ribeiro-etal-2016-trust} proposed Local Interpretable Model-agnostic Explanations (LIME) to explain the predictions of any classifier in an interpretable and faithful manner by learning an interpretable model locally around the prediction. \cite{NIPS2017_7062} proposed the SHapley Additive exPlanations (SHAP) method to assign each feature an importance value for a particular prediction. Although they perform effectively on basic NLP tasks such as sentiment analysis~\cite{hwang-lee-2021-semi}, they do not provide a convincing explanation to users in ChatGPT detection task \cite{mitrović2023chatgpt}, compelling us to step forward and address this gap.

\section{Method}
\label{method}
\begin{figure}[ht]
\includegraphics[width=1\textwidth]{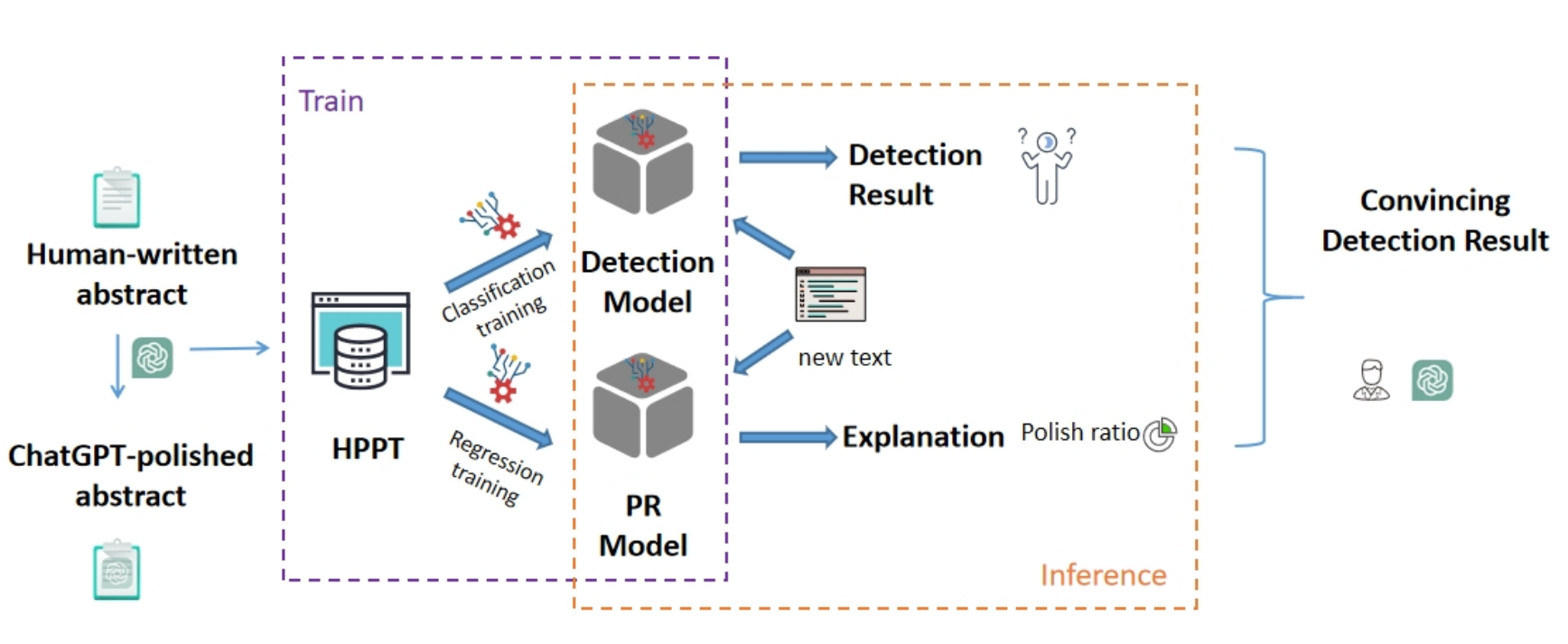}
\caption{The study design of our detection method.}
\label{process}
\end{figure}

To facilitate detecting ChatGPT-polished texts and offer more intuitive explanations to assist final judgment, we first collect human-written abstracts and polish all of them using ChatGPT forming Human-ChatGPT Polished Paired abstracT (HPPT) dataset. Additionally, for each abstract pair in the dataset, we furnish three different similarity metrics (Jaccard Distance, Levenshtein Distance, and BERT semantic similarity) between the human-written abstract and the corresponding abstract polished by ChatGPT. Based on the data, we train the Roberta model as the detector to conduct the detection task. Meanwhile, we also train the Polish Ratio model to explain the detection result to reveal the degree of ChatGPT involvement. The overall detection process is shown in Figure \ref{process}.

\subsection{HPPT Dataset: Human-ChatGPT Polished Paired abstractT}

Since the existing datasets are specifically designed for purely ChatGPT-generated detection, such as HC3 \cite{guo2023close} and ignore the ChatGPT-polished texts, we first construct the Human-ChatGPT Polished Paired abstracT (HPPT) dataset. Specifically, we collect human-written abstracts of accepted papers from several popular NLP academic conferences and polish all of them using ChatGPT~\footnote{The prompt is "please polish the following sentences:<abstracts>". We also tested the prompt "please rewrite the following sentences:<abstracts>" and found that there is no big difference using "polish" or "rewrite". The differences of Levenshtein Distance or Jaccard Distance between using “polish” and “rewrite” for most sample pairs are within the range of $0.1$.}. The texts in our dataset are paired, making it easy to observe the difference between human-written and ChatGPT-polished texts. Overall, we collect 6050 pairs of abstracts and corresponding polished versions from the ACL anthology (ACL, EMNLP, COLING, and NAACL) in the past five years (2018-2022): 2525 are from ACL, 914 are from EMNLP, 1572 are from COLING,  and 1039 are from NAACL. 

\begin{figure}[h]
    \centering
    \begin{subfigure}{0.48\textwidth}
        \centering
        \includegraphics[width=\textwidth]{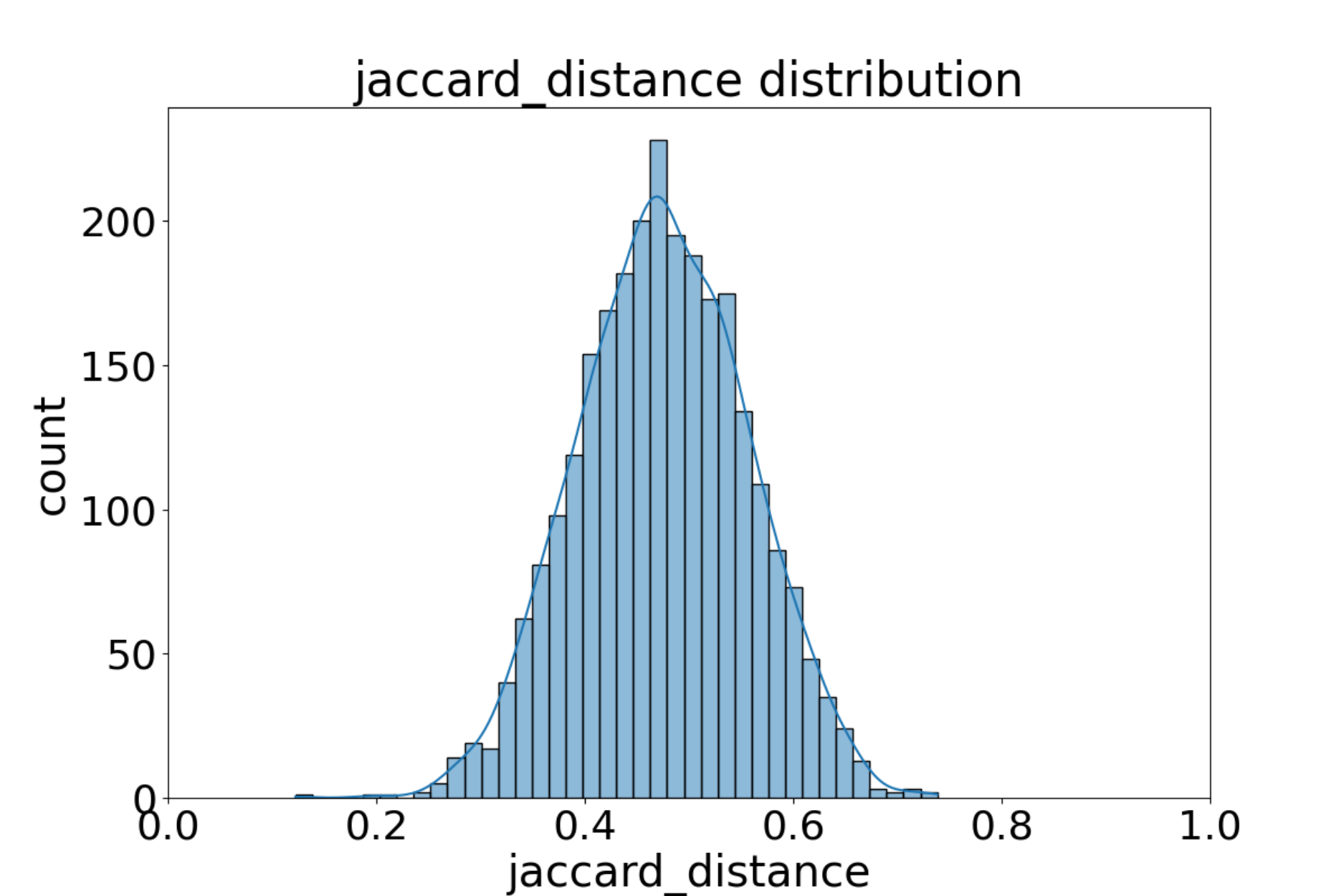}
        \caption*{Jaccard Distance}
    \end{subfigure}
    \hfill 
    \begin{subfigure}{0.48\textwidth}
        \centering
        \includegraphics[width=\textwidth]{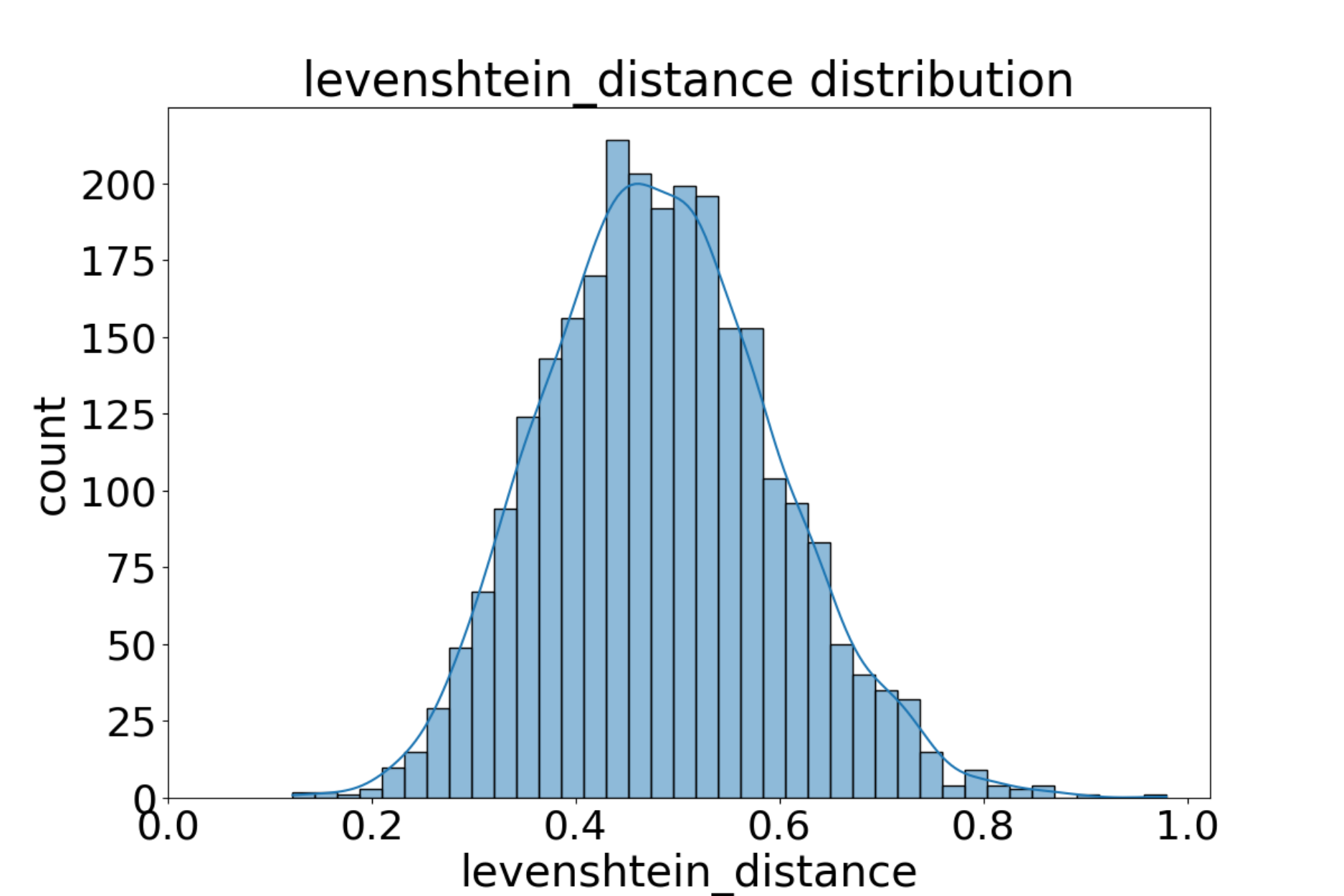}
        \caption*{Levenshtein Distance}
    \end{subfigure}

    \begin{subfigure}{0.48\textwidth}
        \centering
        \includegraphics[width=\textwidth]{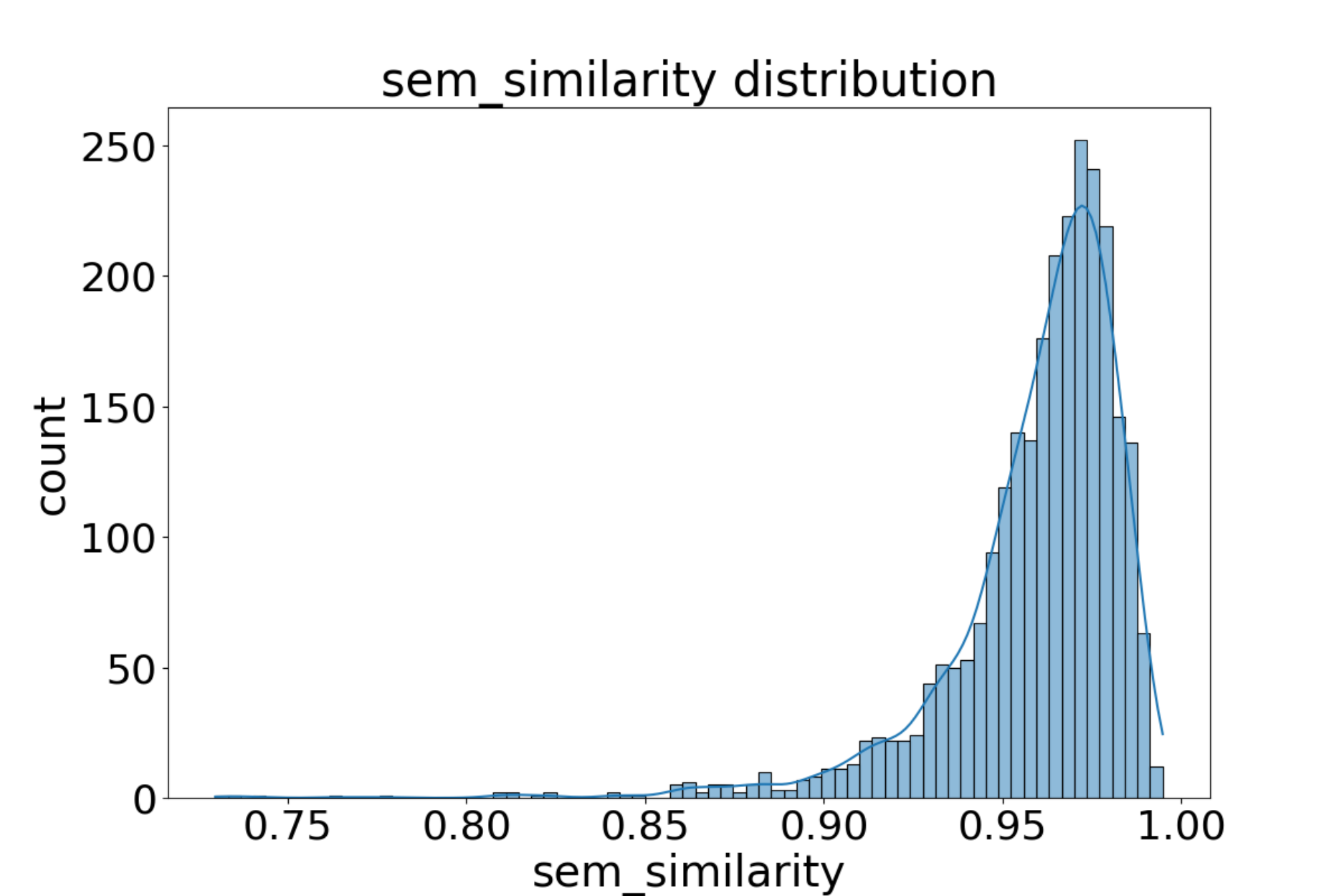}
        \caption*{BERT semantic similarity}
    \end{subfigure}
    
    \caption{Similarity distribution of polished abstracts in our HPPT dataset.}
    \label{similarity}
\end{figure}

To uncover the distinctions between human-written and ChatGPT-polished texts, we compute their similarities using three metrics: BERT semantic similarity\footnote{BERT semantic similarity refers to the cosine similarity between two sentences' embeddings using the BERT model. Here we use the SciBERT model \cite{beltagy-etal-2019-scibert}.}, Jaccard Distance, and Levenshtein Distance (details are presented in section \ref{explanation_method}). As shown in Figure \ref{similarity}, the paired texts are semantically similar in BERT semantic similarity, which shows it is difficult to distinguish human-written and ChatGPT-Polished abstracts from BERT-based semantics. But Jaccard Distance and Levenshtein Distance provide a better way to distinguish them as they align closely with the Gaussian distribution, making them suitable for measuring the degree of ChatGPT involvement.

\subsection{Detection: Roberta-based black box model}

Following the previous work~\cite{guo2023close}, we treat the detection as a binary classification task and build a black-box detector utilizing the Roberta model \cite{liu2019roberta} because the model of ChatGPT is not accessible, as discussed in the Section \ref{rw}. Unlike them, we regard the original abstract without polishing as human-written, and its corresponding ChatGPT polished abstract is regarded as ChatGPT involved.

\subsection{Explanation}
\label{explanation_method}

Due to the black box nature of many existing detectors, they are unable to provide explanations that are crucial for differentiating ChatGPT-Polished text. Therefore, we employ two independent explanation methods GLTR (Giant Language model Test Room) and Polish Ratio (PR). This approach is favored over posterior explanation methods such as SHAP, ensuring more comprehensive justifications for our final judgments.

\begin{itemize}
    \item[$\bullet$]\textbf{GLTR: Giant Language model Test Room} \cite{gehrmann-etal-2019-gltr}  is a statistical method used for AI-generated text detection (mainly for GPT-2). It assumes that human-written text uses a wider subset of the distribution under a model. Although it was initially designed for GPT-2-generated text detection, we formulate a hypothesis that the distribution of GPT-2 and ChatGPT-generated texts is similar in some way since both are AI-generated texts. It uses three criteria to help detection: (1) The probability of the word, as shown in Equation~\ref{prob}; (2) The absolute rank of a word that is the rank of the Equation \ref{prob}; (3) The entropy of the predicted distribution, as shown in Equation~\ref{entropy}.

    \begin{equation}
    \label{prob}
    Prob_i = p_{det}(X_i|X_{1:i-1})
    \end{equation}
    
    \begin{equation}
    \label{entropy}
        Entropy_i = -\sum_w p_{det}(X_i = w |X_{1:i-1})log(\\p_{det}(X_i = w |X_{1:i-1}))
    \end{equation}

    Specifically, if a text is written by a human, a word should have a low probability, which leads to a higher top rank and the entropy also should be large.

    We consider the GLTR as our baseline for the explanation method as we have discovered that the method is effective in explaining the distinction between human-written and entirely ChatGPT-generated texts. However, it fails to provide adequate explanations when ChatGPT-polished texts are present (refer to Section \ref{Explanation Analysis} for further details). To address this issue, we adopt a new method called the Polish Ratio.
\end{itemize}
\begin{itemize}
    \item[$\bullet$]\textbf{PR: Polish Ratio} is a new metric we propose to measure the degree of ChatGPT involvement for a text. It measures the degree of text modification before and after revision using ChatGPT by calculating the similarity score between original human-written texts and the corresponding polished texts. In our dataset HPPT, we take two metrics \textbf{Jaccard Distance} and \textbf{Levenshtein Distance} (Levenshtein Distance is normalized by the maximum length of the two sequences) as the Polish Ratio. Since ChatGPT generates the text using tokens, we evaluate the distances based on white-space tokenized words. As shown in Equation~\ref{jaccard}, Jaccard distance measures the dissimilarity between sets by comparing the size of their intersection and union. Levenshtein distance measures the difference between two sequences by counting the minimum number of edits (deletion, insertion, and substitution), as shown in Equation~\ref{lev}.

\begin{equation}
\label{jaccard}
J(A, B) = 1 - \frac{{|A \cap B|}}{{|A \cup B|}}
\end{equation}

Where:
\begin{itemize}
  \item \(A\), \(B\) are the set of words in the document text strings.
  \item \(|\cdot|\) represents the cardinality (number of elements) of the set.
\end{itemize}

\begin{equation}
\label{lev}
\text{lev}(a, b) = 
\begin{cases}
    \max(|a|, |b|)\ \  \text{if } \min(|a|, |b|) = 0 \\
    \min \left\{
    \begin{aligned}
    & \text{lev}(a[2:], b[2:]) + 1, \\
    & \text{lev}(a[2:], b[1:]) + 1, \\
    & \text{lev}(a[1:], b[2:]) + 1
    \end{aligned}
    \right\} \text{otherwise}
\end{cases}
\end{equation}

Where:
\begin{itemize}
    \item \(a\), \(b\) represent a list of white-space tokenized words in the document.
    \item \(|\cdot|\) represents the length of the word list.
    \item \(\text{lev}(a[i:], b[j:])\) represents the Levenshtein distance between the sublist of \(a\) starting at index \(i\) and \(b\) starting at index \(j\).
\end{itemize}

 Therefore, we regard the PR model as the regression model where either the Jaccard distance or normalized Levenshtein distance of the polished texts is the target value of the Polish Ratio. In ideal conditions, the predicted PR value of an abstract should approach 0 for a human-written one and should be close to 1 when ChatGPT revises a majority of words in the abstract. We use the Roberta-based model to extract feature $\mathbf{x_i}\in \mathbb{R}^{768}$ of each sentence $i$ and use an MLP to conduct the final regression task. The sigmoid function is chosen as the activation function to ensure the range of the regression result is in $[0,1]$. Compared to other explanation indices like confidence level, our PR method takes advantage of the paired abstracts before and after polishing to measure how much the ChatGPT involvement is, which can give a more independent and convincing explanation.
\end{itemize}

\section{Experiment and analysis}
\subsection{Experiment Setup}
\subsubsection{Dataset}
We conduct experiments on the following three datasets to demonstrate the effectiveness of our model.

\begin{itemize}
    \item HPPT
    
    It is our built ChatGPT-polished dataset, which consists of 6050 pairs of recent abstracts and corresponding polished versions. Meanwhile, to measure the degree of ChatGPT involvement in the text, we also provide the Levenshtein distance and Jaccard distance of the polished abstracts compared with their corresponding human-written ones as the labeled PR value and label 0 as the PR value of those human-written abstracts. We randomly partition the HPPT into the train, test, and validation sets by $6:3:1$ to train and test our model (Roberta-HPPT).
    
   \item HC3 \cite{guo2023close}

    The HC3 (Human ChatGPT Comparison Corpus) dataset\footnote{\url{https://huggingface.co/datasets/Hello-SimpleAI/HC3/tree/main}} is one of the most popular ChatGPT detection datasets containing question-answering QA pairs. The human answers are collected from publicly available question-answering datasets and wiki text, while the answers provided by ChatGPT are obtained from its preview website through manual input of questions for each interaction. We choose its English version corpus, which consists of 85,449 QA pairs (24,322 questions, 58,546 human answers, and 26,903 ChatGPT answers). We randomly partition the HC3 into the train, test, and validation sets by $6:3:1$ and regard the answer text as the input of our detection model to ensure the detector's versatility. In addition, it is also used to train our baseline model (Roberta-HC3).

\item CDB \cite{liang2023gpt}

The ChatGPT-Detector-Bias Dataset\footnote{\url{https://huggingface.co/datasets/WxWx/ChatGPT-Detector-Bias}} (CDB) is a mixed dataset consisting of  749 text samples derived from human, ChatGPT and GPT-4. The human data is from TOEFL essays in a Chinese educational forum, US College Admission essays, and scientific abstracts from Stanford’s course CS224n. The GPT data consists of samples polished by both ChatGPT and GPT-4, as well as essays fully generated by ChatGPT using different prompts. We also take it as the harder dataset to test the detectors' generalization ability because it not only contains the GPT-4-generated or GPT-4-polished text but also contains well-designed prompt engineering ChatGPT-generated text and the human writing samples from both native and non-native English writers.

\end{itemize}

\subsubsection{Reproduction details}
\label{Reproduction details}
    During training our detection model, we use the batch size $16$, learning rate $2e-5$ and the maximum epoch is set to 10. The model is chosen as the best one on the validation set. While training our PR model, we use batch size $4$, learning rate $1e-5$, and the maximum epoch is set to 15.

\subsection{Detection Result}

Table \ref{detection} shows several baselines and our model performance on three datasets. Our model performs well on both in-domain dataset HPPT and out-of-domain datasets (HC3 and CDB), suggesting that our model trained on the polished HPPT dataset is more robust than other models. Although the Roberta-HPPT model is only trained on HPPT, it achieves comparable performance compared to the SOTA model in HC3, with only a 3\% difference and better than DetectGPT. On the more difficult dataset (CBD), our model significantly outperforms other baselines with 88.15\% accuracy. Specifically, our model only drops 6\% on the out-of-domain dataset while the Roberta-HC3 and the DetectGPT drop by nearly 40\%, demonstrating the strong robustness of our model. The reason is that our model is trained on the ChatGPT-polished text instead of ChatGPT-generated text, which can tackle more difficult samples such as GPT-4-generated, GPT-4-polished, and well-designed prompt engineering ChatGPT-generated texts. 

\begin{table}[ht]
\footnotesize
\centering
\begin{tabular}{c|ccc|ccc}
\hline
\multirow{2}{*}{\begin{tabular}[c]{@{}c@{}}Test \\ Dataset\end{tabular}} &
  \multicolumn{3}{c|}{ACC} &
  \multicolumn{3}{c}{AUROC} \\ \cline{2-7} 
                                                       & HPPT   & HC3    & CDB    & HPPT   & HC3    & CDB    \\ \hline
GPTZero                                                & -      & -      & 0.4406 & -      & -      & 0.6818 \\ \hline
\begin{tabular}[c]{@{}c@{}}OpenAI-GPT-2\end{tabular} & -      & -      & 0.4633 & -      & -      & 0.5604 \\ \hline
OriginalityAI                                          & -      & -      & 0.4967 & -      & -      & 0.5721 \\ \hline
DetectGPT                                              & 0.5129 & 0.8309 & 0.4593 & 0.6876 & 0.9058 & 0.7308 \\ \hline
\begin{tabular}[c]{@{}c@{}}Roberta-HC3\end{tabular} &
  0.5285 &
  \textbf{0.9991} &
  0.5848 &
  0.7946 &
  \textbf{1} &
  0.7526 \\ \hline
\begin{tabular}[c]{@{}c@{}}Roberta\\ -HPPT (ours)\end{tabular} &
  \textbf{0.9465} &
  0.9671 &
  \textbf{0.8825} &
  \textbf{0.9947} &
  0.9931 &
  \textbf{0.9518} \\ \hline
\end{tabular}
\caption{The detection performance of some popular detection models. The evaluation metrics are Area under ROC curve (AUROC) and accuracy (ACC). The results of GPTZero, OpenAl-GPT-2 \cite{solaiman2019release}, and OriginalityAI detectors are derived from the data presented in \cite{liang2023gpt}.}
\label{detection}
\end{table}

\subsection{Explanation Analysis}
\label{Explanation Analysis}

Although our model achieves high accuracy in detecting ChatGPT-polished texts, it still needs an explanation of the degree of ChatGPT involvement in the text. Motivated by this, we adopt two explanation methods (GLTR and Polish Ratio) to measure them. 

\subsubsection{GLTR}
\begin{figure}[ht]
\centering
    \begin{subfigure}[ht]{0.495\textwidth}
        \includegraphics[width=\textwidth]{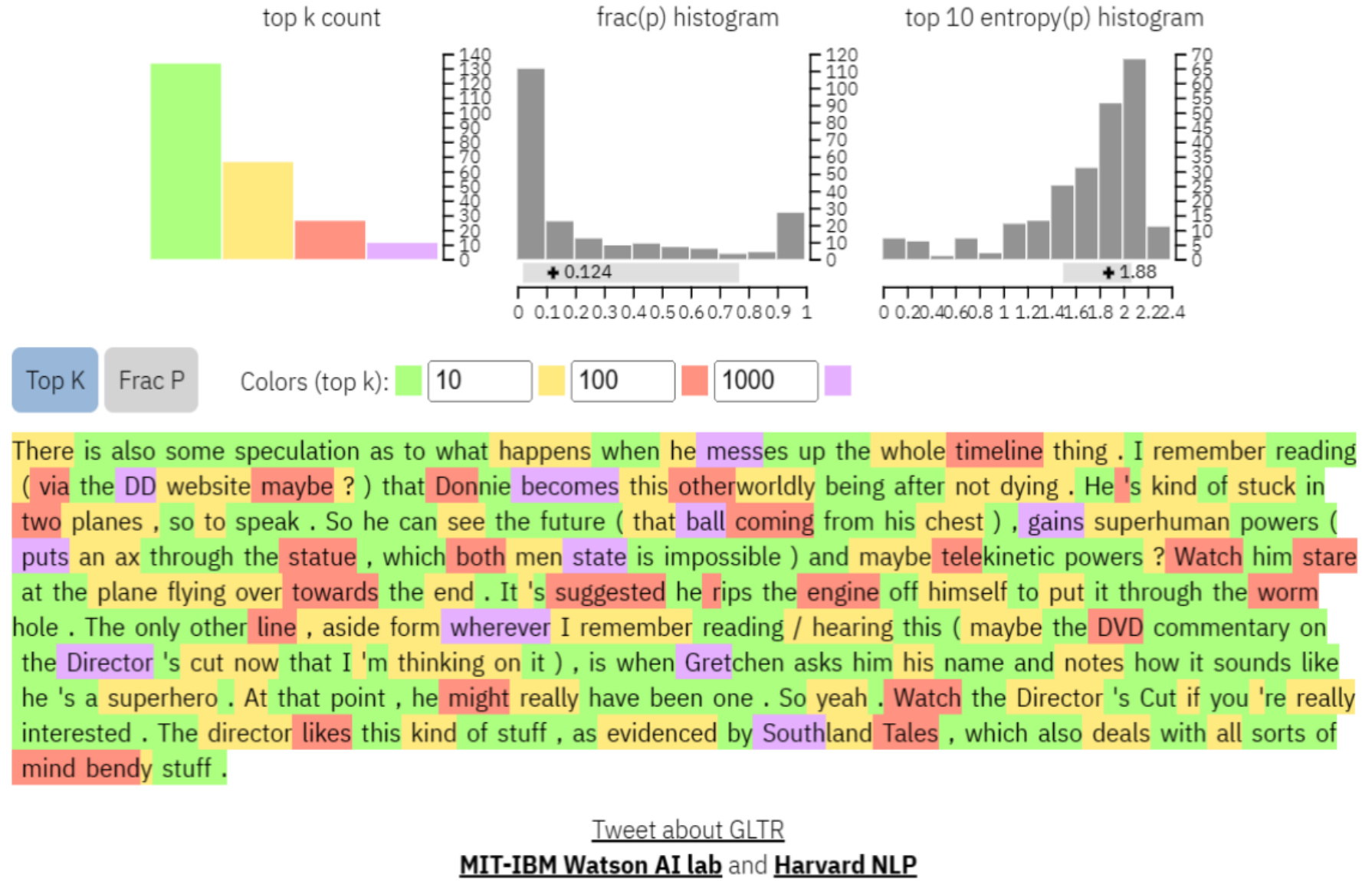}
        \caption{Sample 1: human-written}
        \label{a}
    \end{subfigure}
    \begin{subfigure}[ht]{0.495\textwidth}
        \includegraphics[width=\textwidth]{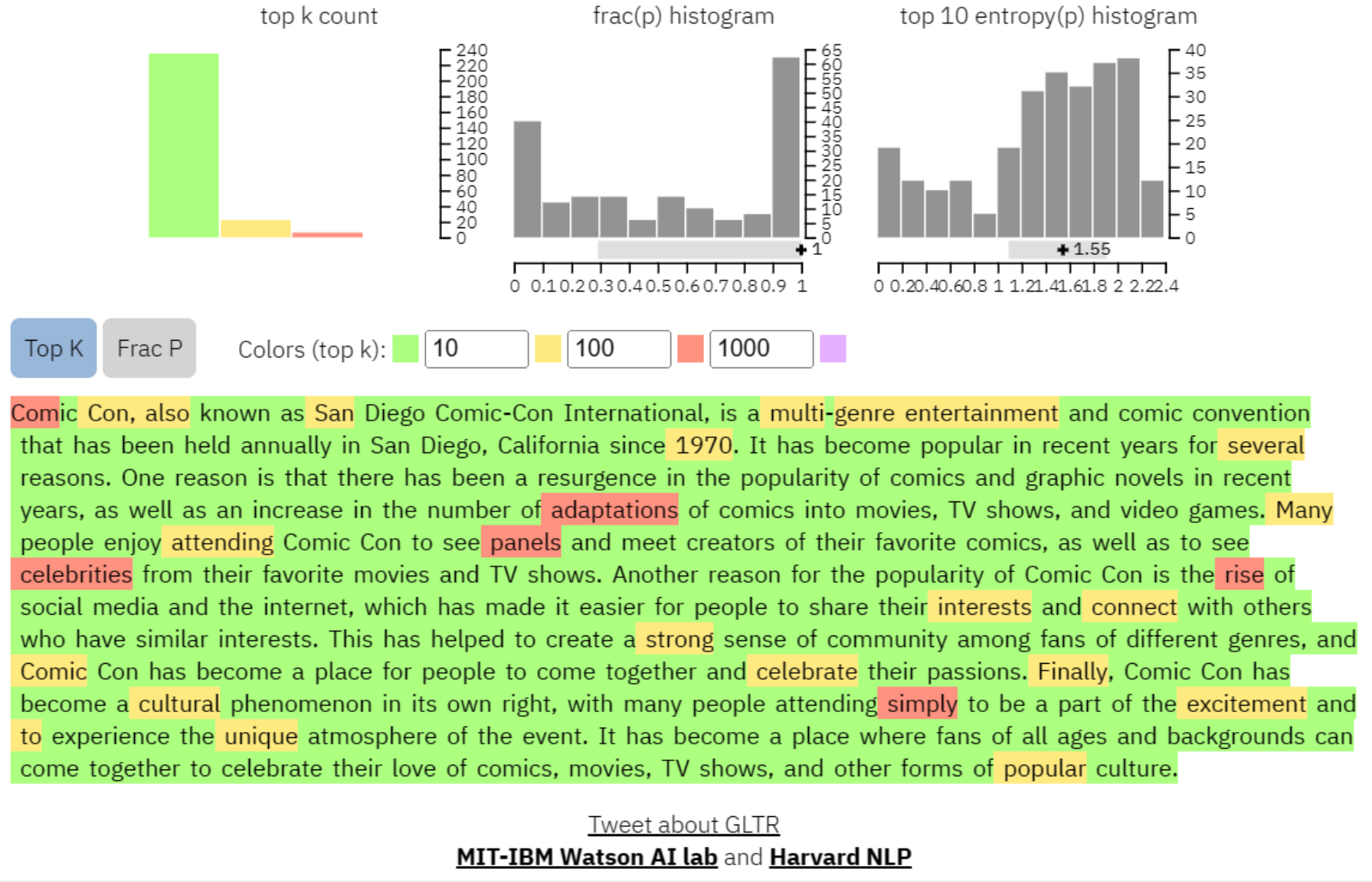}
        \caption{Sample 2: ChatGPT-generated}
        \label{b}
    \end{subfigure}
    \\
    \begin{subfigure}[ht]{0.495\textwidth}
        \includegraphics[width=\textwidth]{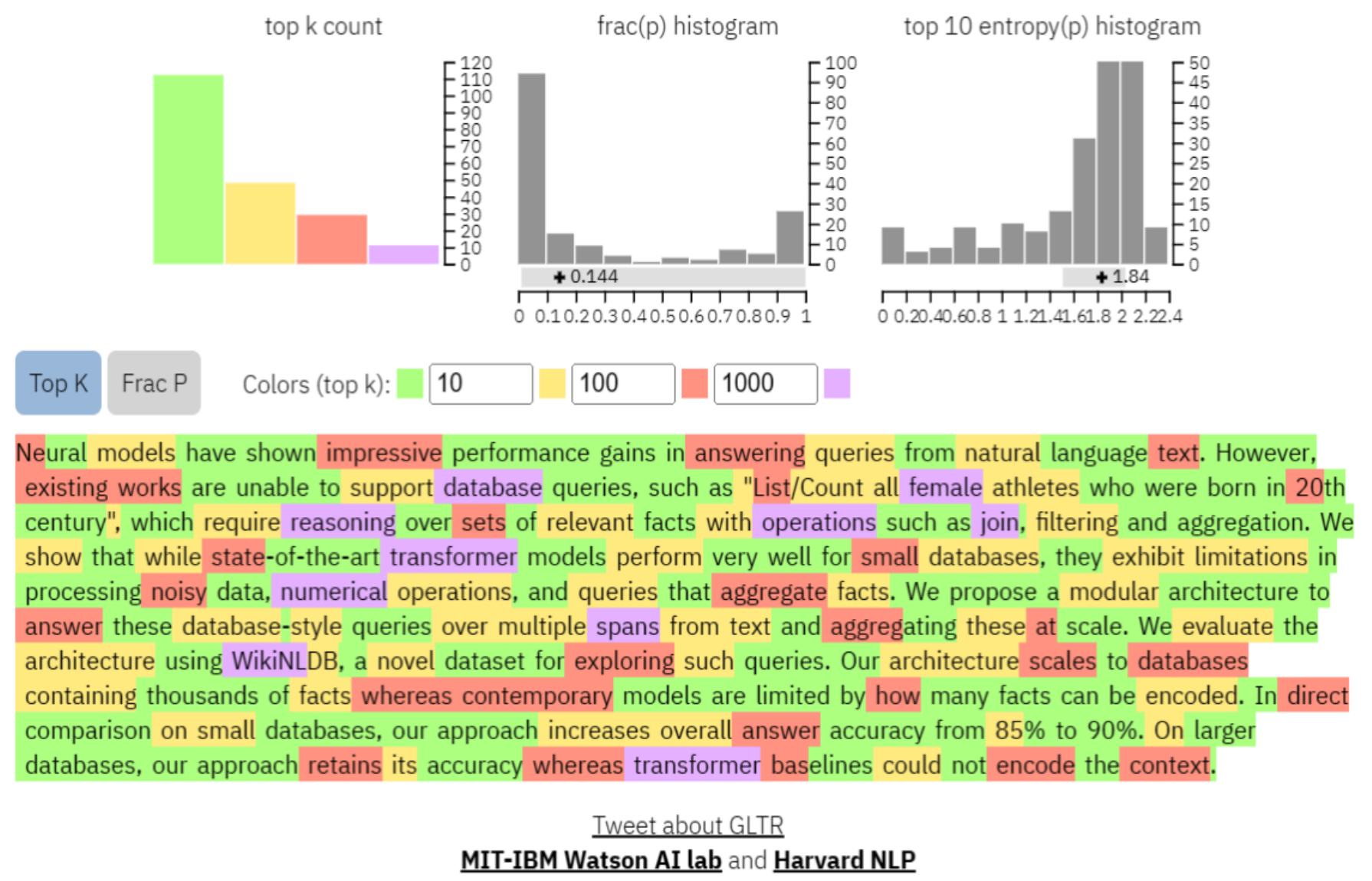}
        \caption{Sample 3: human-written}
        \label{d}
    \end{subfigure}
    \begin{subfigure}[ht]{0.495\textwidth}
        \includegraphics[width=\textwidth]{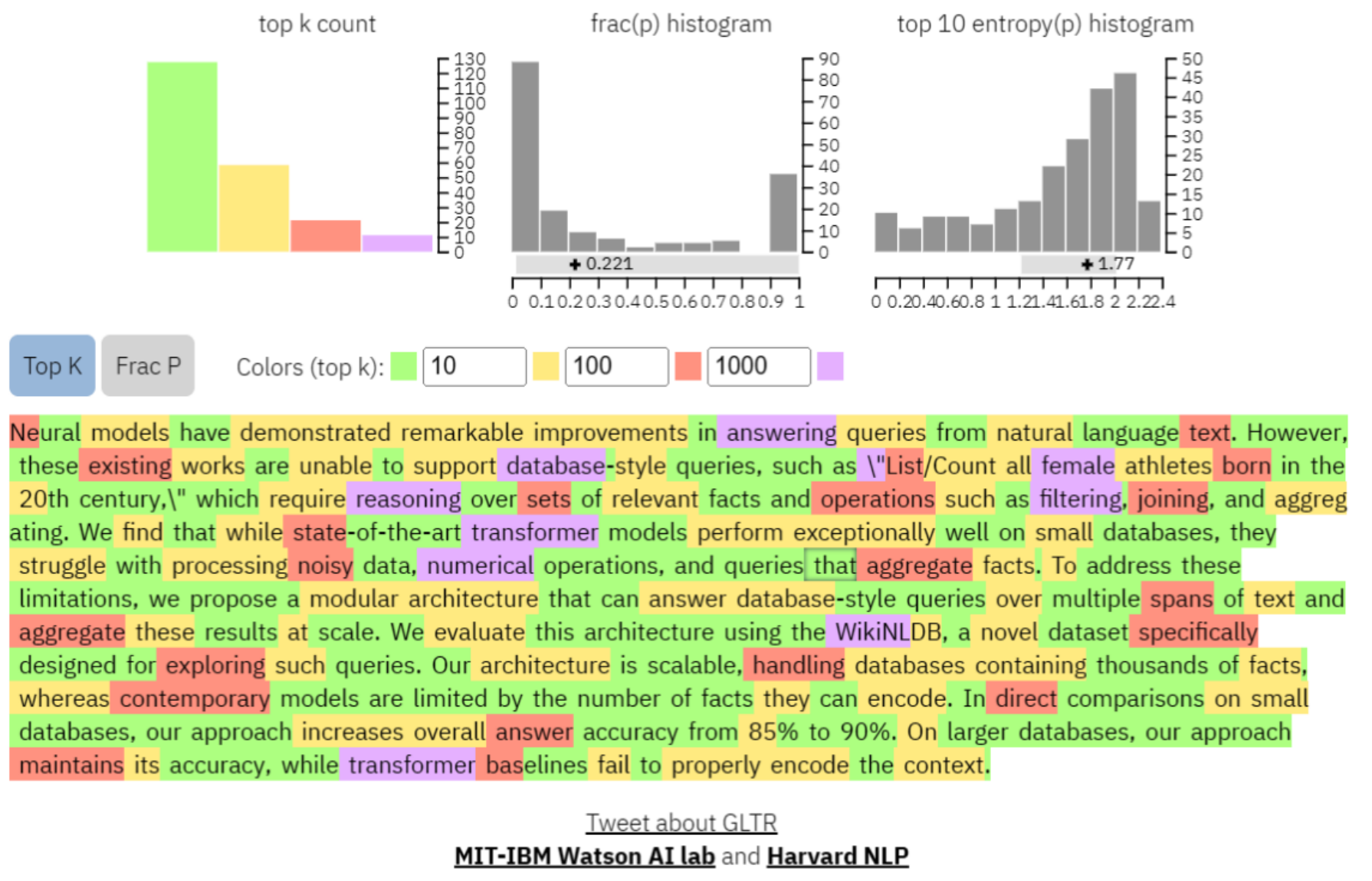}
        \caption{Sample 4: ChatGPT-polished}
        \label{c}
    \end{subfigure}
    \\
\caption{The visualization result of some sample texts with the help of the GLTR demo: \url{http://gltr.io./dist/index.html}. A word that ranks within the top 10 probability is highlighted in green, top 100 in yellow, top 1,000 in red, and the rest in purple. Sample 1 and Sample 2 are chosen from the HC3 test set, while Sample 3 and Sample 4 are chosen from the HPPT test set.}
\label{GLTR}
\end{figure}


Figure \ref{GLTR} shows the visualization of the probability, absolute rank, and the distribution's entropy of two pairs of texts from HC3 and HPPT.  A higher concentration of green tokens indicates that the words are selected from the top-ranked probability generated by the LLM system. Comparing samples 1 and 2, we find that the difference between human-written and ChatGPT-generated texts is noticeable. Texts entirely generated by ChatGPT prefer to use those top-rank probability words, which validates our hypothesis in Section \ref{explanation_method}.  However, it is relatively hard to distinguish human-written and ChatGPT-polished texts, as illustrated in samples 3 and 4. The reason is that the fraction of top-rank probability words employed in polished texts decreases significantly, approaching that of human-written texts. As a result, the GLTR explanation mechanism proves challenging in this scenario.

\subsubsection{Polish Ratio Regression}

Therefore, we propose Polish Ratio to explain the detection result of both entirely ChatGPT-generated and ChatGPT-polished texts. As shown in Table~\ref{PR}, we try Jaccard distance and Levenshtein distance as the true label of the Polish Ratio respectively. Considering the influence of outliers and the unbalanced samples because of 0 PR ones (human-written abstracts), we try different loss functions to test their effectiveness. Suppose the target value for the Polish Ratio of each text sample $i$ is $y_i$ and the predicted value is $\hat{y}_i$, the different losses $\ell_i$  for sample $i$ are defined as:
\begin{equation}
    \mathcal{L}_1(y_i, \hat{y_i}) = |y_i - \hat{y}_i|
\end{equation}
\begin{equation}
    Smooth\mathcal{L}_1(y_i, \hat{y_i}) = \begin{cases}
0.5 \cdot (y_i - \hat{y_i})^2 & \text{if } |y_i - \hat{y_i}| < \beta \\
|y_i - \hat{y_i}| - 0.5 \cdot \beta & \text{otherwise}
\end{cases}
\end{equation}
\begin{equation}
    \mathcal{L}_{MSE}(y_i, \hat{y_i}) = (y_i - \hat{y}_i)^2
\end{equation}
where $\mathcal{L}_1$ is L1-loss, $Smooth\mathcal{L}_1$ is smooth L1-loss and $\mathcal{L}_{MSE}$ is the mean square error while the loss for all $N$ samples is calculated as $\ell(y,\hat{y}) = \frac{1}{N}\sum_{i=1}^N \ell_i$. Through the experiment, we find that MSE is the best choice. 

\begin{table}[t]
\small
\centering
\begin{tabular}{ccc}
\hline
                         & \multicolumn{2}{c}{\textbf{Similarity metrics}} \\ \hline
\textbf{Loss function}   & Jaccard distance     & Levenshtein distance     \\ \hline
L1-loss                  & 0.0837               & 0.0832                   \\ \hline
Smooth L1-loss ($\beta=0.1$) & 0.0779               & 0.0866                   \\ \hline
Mean Square Error         & \textbf{0.0728}      & \textbf{0.0813}          \\ \hline
\end{tabular}
\caption{Comparison of mean absolute error (MAE) values for PR regression using different loss functions and similarity metrics on the HPPT test set. Lower values indicate better performance.}
\label{PR}
\end{table}
\begin{figure}[ht]
\centering
    \begin{subfigure}[t]{0.475\textwidth}
    \includegraphics[width=\textwidth]{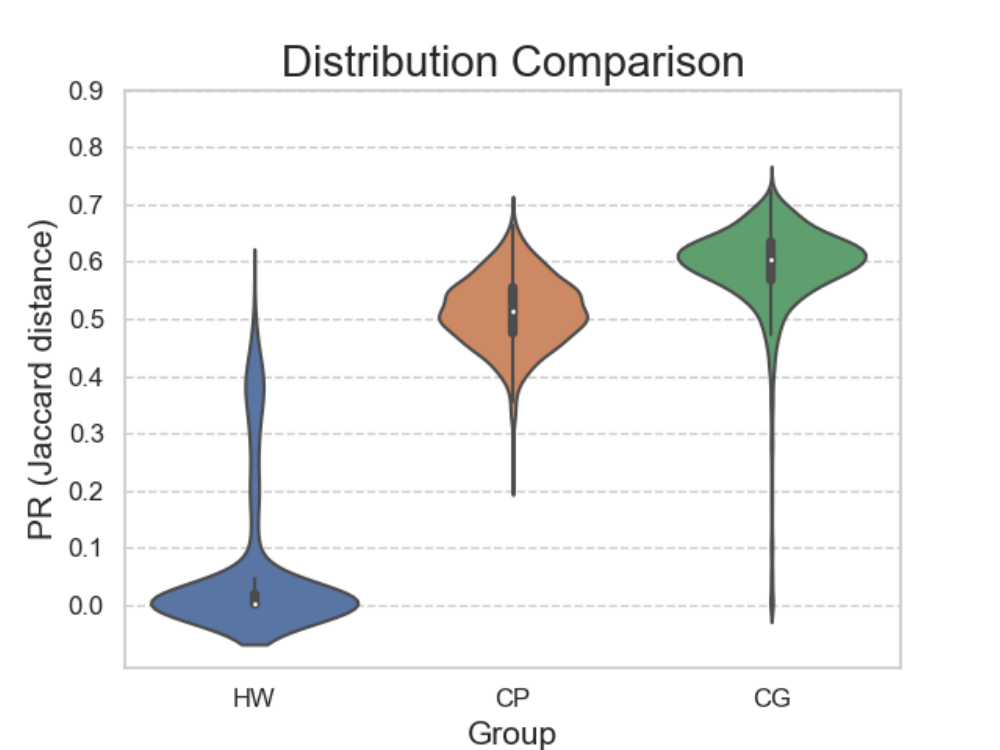}
    \caption{Jaccard distance}
    \end{subfigure}
    \begin{subfigure}[t]{0.475\textwidth}
    \includegraphics[width=\textwidth]
   {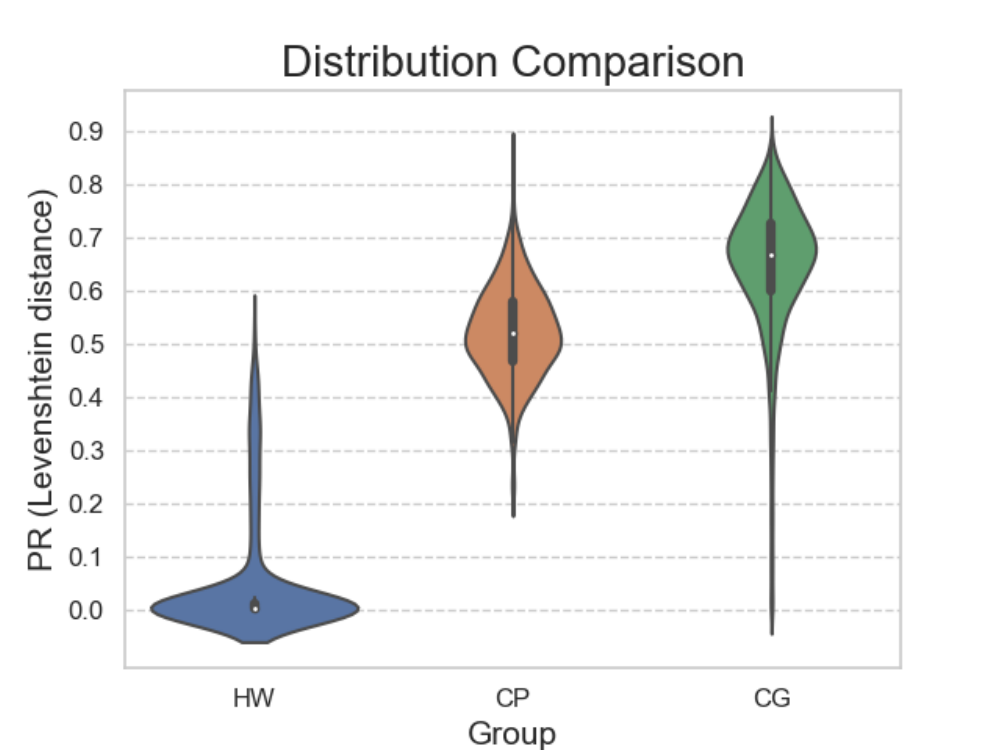}
    \caption{Levenshtein distance}
    \label{PR_lev}
    \end{subfigure}

\caption{Differences between predicted PR for human-written texts (HW), ChatGPT-polished texts (CP) and ChatGPT-generated texts (CG): HW and CP are directly from the HPPT testset where CG are from the HC3 testset.}
\label{PR_dif}
\end{figure}

\begin{table}[]
\centering
\tiny
\begin{tabular}{|p{8cm}|p{1cm}|p{0.8cm}|}
\hline
\textbf{Text} & \textbf{Source} & \textbf{PR} \\
\hline
1a) Theoretical work in morphological typology offers the possibility of measuring morphological diversity on a continuous scale. However, literature in Natural Language Processing (NLP) typically labels a whole language with a strict type of morphology, e.g. fusional or agglutinative. In this work, we propose to reduce the rigidity of such claims, by quantifying morphological typology at the word and segment level. We consider Payne (2017)'s approach to classify morphology using two indices: synthesis (e.g. analytic to polysynthetic) and fusion (agglutinative to fusional). For computing synthesis, we test unsupervised and supervised morphological segmentation methods for English, German and Turkish, whereas for fusion, we propose a semi-automatic method using Spanish as a case study. Then, we analyse the relationship between machine translation quality and the degree of synthesis and fusion at word (nouns and verbs for English-Turkish, and verbs in English-Spanish) and segment level (previous language pairs plus English-German in both directions). We complement the word-level analysis with human evaluation, and overall, we observe a consistent impact of both indexes on machine translation quality. & Human & 0.14\% \\
\hline
1b) Theoretical work in morphological typology \textcolor{red}{provides a means of quantifying} morphological diversity on a continuous scale. However, NLP literature often labels a language with a strict morphological type, \textcolor{red}{such as} fusional or agglutinative. \textcolor{red}{This work aims} to reduce the rigidity of these claims by quantifying morphological typology at \textcolor{red}{both} the word and segment level. We \textcolor{red}{adopt} Payne (2017)'s approach \textcolor{red}{, which} classifies morphology using two indices: synthesis (e.g. analytic to polysynthetic) and fusion (agglutinative to fusional). \textcolor{red}{To compute} synthesis, we \textcolor{red}{evaluate} unsupervised and supervised morphological segmentation methods for English, German, and Turkish. For fusion, we propose a semi-automatic method using Spanish as a case study. We \textcolor{red}{examine} the relationship between machine translation quality and synthesis and fusion at \textcolor{red}{both} the word level (nouns and verbs for English-Turkish and verbs in English-Spanish) and segment level \textcolor{red}{(English-German in both directions).} We \textcolor{red}{supplement} the word-level analysis with human evaluation, and we observe a consistent impact of both indices on machine translation quality.
 & ChatGPT-polished & 34.48\% \\
\hline
\hline
2a) Framing is a communication strategy to bias discussion by selecting and emphasizing. Frame detection aims to automatically analyze framing strategy. Previous works on frame detection mainly focus on a single scenario or issue, ignoring the special characteristics of frame detection that new events emerge continuously and policy agenda changes dynamically. To better deal with various context and frame typologies across different issues, we propose a two-stage adaptation framework. In the framing domain adaptation from pre-training stage, we design two tasks based on pivots and prompts to learn a transferable encoder, verbalizer, and prompts.In the downstream scenario generalization stage, the transferable components are applied to new issues and label sets. Experiment results demonstrate the effectiveness of our framework in different scenarios. Also, it shows superiority both in full-resource and low-resource conditions. & Human & 0.14\% \\
\hline
2b) \textcolor{red}{The communication strategy of framing involves selecting and emphasizing certain aspects in order to bias discussion. To analyze this strategy, frame detection is used. However,} previous works \textcolor{red}{in this field} have mainly focused on a single scenario or issue, ignoring the \textcolor{red}{fact that new events and policy agendas are constantly emerging. To address this issue,} we propose a two-stage adaptation framework. \textcolor{red}{The first stage involves adapting the framing domain through pre-training, using} two tasks based on pivots and prompts to learn a transferable encoder, verbalizer, and prompts. \textcolor{red}{In the second stage,} the transferable components are applied to new issues and label sets. Our framework \textcolor{red}{has been shown to be effective} in different scenarios, \textcolor{red}{and to perform better than other methods,} both in conditions of full resources and low resources. & ChatGPT-polished & 66.61\% \\
\hline
\end{tabular}
\caption{Sample cases of the Polish Ratio based on the Levenshtein distance where the parts edited by ChatGPT are highlighted in red. From a subjective standpoint, we can discern the rationale behind our PR regression task: the PR score for human-written abstracts tends to be close to 0, indicating no modification. Also, larger PR means more parts introduced by ChatGPT as represented by more red parts.}
\label{PR_case}
\end{table}

The different distributions of PR values given by our model on test data in Figure \ref{PR_dif} show that our PR explanation model demonstrates strong generalization capabilities. It not only effectively distinguishes between human-written and polished texts but also successfully discriminates ChatGPT-generated texts that are not encountered during the training stage. While both Jaccard distance and Levenshtein distance serve as suitable PR metrics to differentiate the CP and CG groups, the use of Levenshtein distance proves to be more effective, where the median Polish Ratio of ChatGPT-polished texts is around 0.5, while that of ChatGPT-generated texts is around 0.65. According to the violin plot shown in Figure \ref{PR_lev}, we suggest a Polish Ratio greater than 0.2 indicates ChatGPT involvement, and a value greater than 0.6 indicates that ChatGPT generates most words.

\subsection{Case study for Polish Ratio}
Our Polish Ratio is also accurate in the case study. Table \ref{PR_case} is the case study of the Polish Ratio regression on the index Levenshtein Distance. Indeed, it can effectively differentiate the extent of modification introduced by ChatGPT. For instance, the degree of text modification by ChatGPT varies significantly between text (1b) and text (2b), as highlighted in red, leading to a notable difference in their PR values, which proves the effectiveness of our PR method.

\begin{table}[ht]
\centering
\begin{tabular}{ccccc}
\hline
                 & Precision & Recall  & Support & Model                         \\ \hline
\textbf{Human}   & 51.47\%   & 99.46\% & 738     & \multirow{2}{*}{Roberta-HC3}  \\ \cline{1-4}
\textbf{ChatGPT} & 92.00\%   & 6.23\%  & 738     &                               \\ \hline \hline
\textbf{Human}   & 99.4\%    & 89.84\% & 738     & \multirow{2}{*}{Roberta-HPPT} \\ \cline{1-4}
\textbf{ChatGPT} & 90.70\%   & 99.46\% & 738     &                               \\ \hline
\end{tabular}
\caption{Detailed recall and precision for the Roberta-HC3 and Robarta-HPPT in HPPT test Corpus.}
\label{mis-all}
\end{table}

\subsection{Error Analysis}
We conduct an error analysis to gain insights into our model's learning process and assess the efficacy of our explanation methods. We focus on understanding the reasons behind the mistakes the baseline model and our model made in the HPPT dataset, shown in Table \ref{mis-all}. 

\begin{figure}[ht]
\centering
        \includegraphics[scale=0.5]{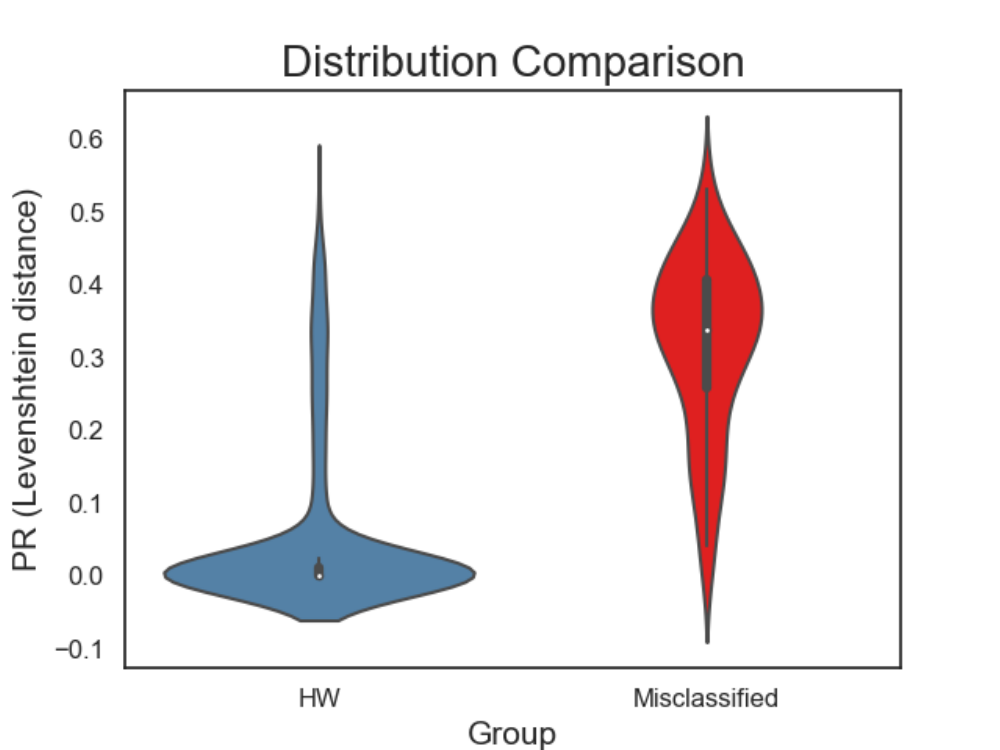}
        \caption{Differences between human-written samples and misclassified samples whose ground truth is human-written in the test set: the mean value of PR for misclassified samples is around 0.3, which makes our detection model confused.}
        \label{comparison_2}
\end{figure}

The result shows that the baseline model only trained on the purely ChatGPT-generated text exhibited confusion when encountering a polishing attack, as evidenced by the classification of nearly all polished samples as human-written. In contrast, our model (Roberta-HPPT) tends to make mistakes for originally human-written texts, as shown in Table~\ref{mis-all}. By exploring the Polish Ratio of these misclassified samples, we find misclassified samples have a relatively high PR which originally should be close to 0 PR for human-written samples, as shown in Figure \ref{comparison_2}. This discovery indicates that our PR method can explain the reasons why the detection model makes mistakes: the writing style of original human-written texts may be similar to that of ChatGPT.

\subsection{Generalization to the Chinese language and different LLMs}
\begin{figure}[]
\centering
    \begin{subfigure}[t]{0.48\textwidth}
    \includegraphics[width=\textwidth]{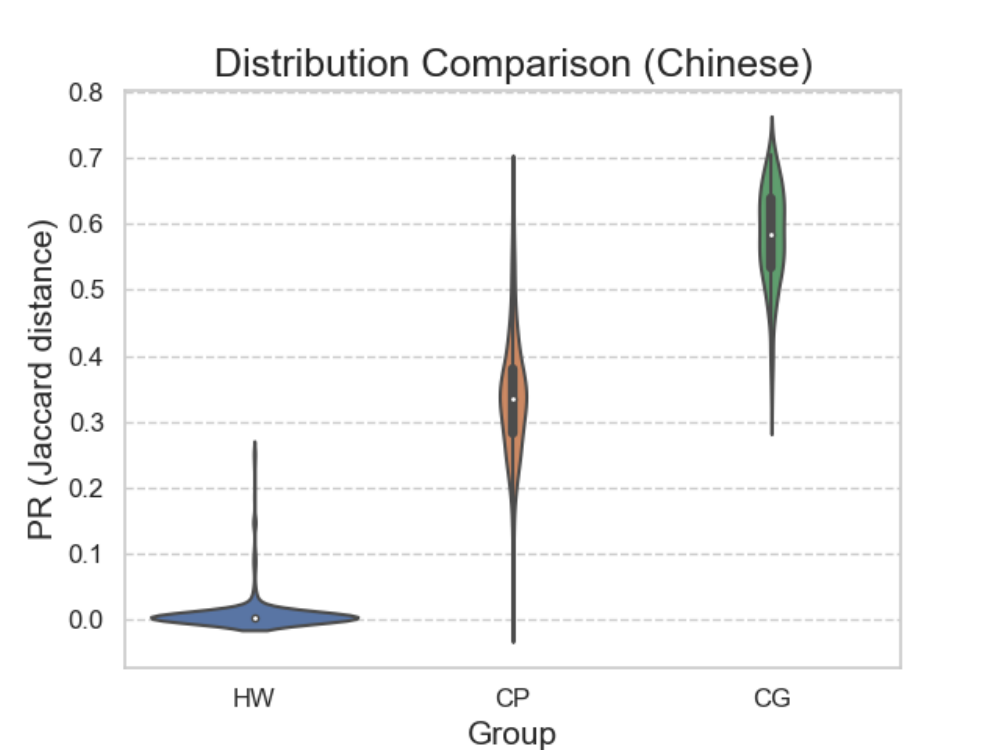}
    \caption{Jaccard distance}
    \end{subfigure}
    \begin{subfigure}[t]{0.48\textwidth}
    \includegraphics[width=\textwidth]{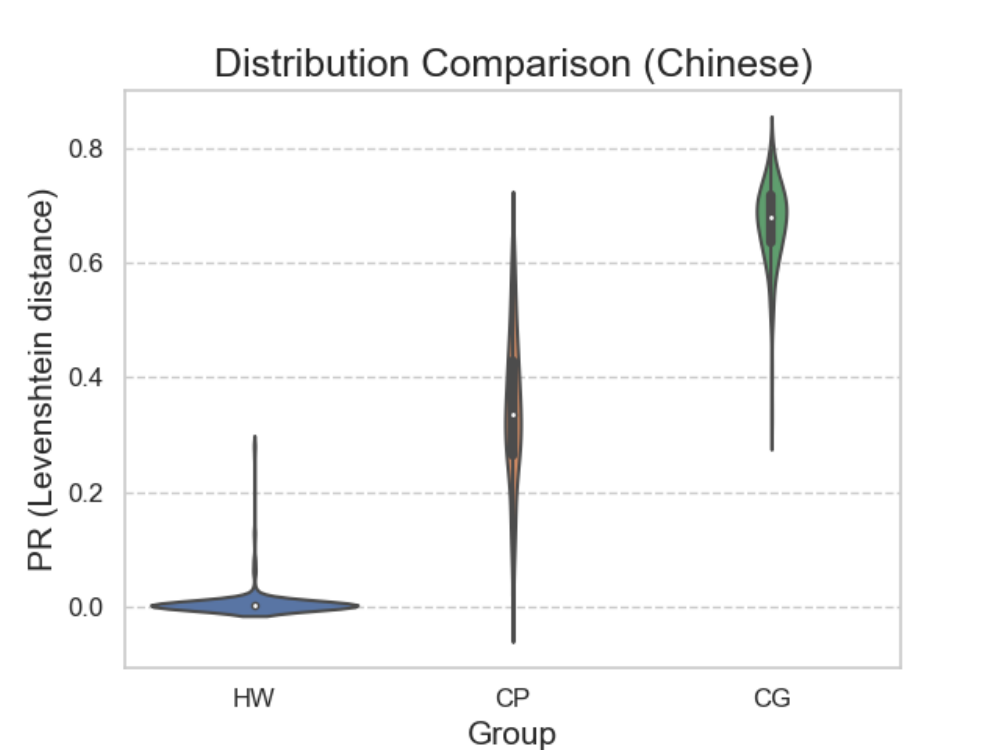}
    \caption{Levenshtein distance}
    \end{subfigure}
    \caption{Differences between predicted PR for human-written texts (HW), ChatGPT-polished texts (CP) and ChatGPT-generated texts (CG).}
    \label{comparison_chi}
\end{figure}

The framework of our ChatGPT involvement detection method also works effectively on Chinese texts. We collect 1,232 abstracts from the CSL: a large-scale Chinese Scientific Literature dataset \cite{li-etal-2022-csl} that contains 396,209 Chinese papers' meta-information including title, abstract, keywords, academic category and discipline. Then we use ChatGPT to polish \footnote{The prompt is "\begin{CJK*}{UTF8}{gbsn}
请润色以下文本：
\end{CJK*}<abstracts>" (Translation: "Please polish the following sentences:<abstracts>").} the selected 1,232 Chinese abstracts and calculate the Jaccard Distance and Levenshtein Distance (normalized) between the polished and the corresponding human-written abstract based on Chinese characters. The 1,232 pairs of abstracts are randomly partitioned into the train, test, and validation sets by $6:3:1$. Following the framework mentioned in Section \ref{method}, we train the detection and PR model \footnote{We use the Chinese pre-trained Roberta model: \url{https://huggingface.co/hfl/chinese-roberta-wwm-ext}} to investigate whether our proposed method is generalizable to the Chinese language. Through experiments, we discover that the detection accuracy ($98.24\%$) significantly outperforms DetectGPT ($53.25\%$) on our collected Chinese test sets, and our PR model can also distinguish human-written, ChatGPT-polished, and ChatGPT-generated 
texts \footnote{Human-written and ChatGPT-polished texts are from our collected test sets. ChatGPT-generated texts are 100 scientific abstracts additionally generated purely by ChatGPT through the prompt: "\begin{CJK*}{UTF8}{gbsn}
请写一篇200字左右科学文献中文摘要
\end{CJK*}
" (Translation: "Please write a Chinese abstract of scientific literature of about 200 words").
}  effectively as shown in Figure \ref{comparison_chi}. 

\begin{figure}[]
\centering
    \begin{subfigure}[t]{0.48\textwidth}
    \includegraphics[width=\textwidth]{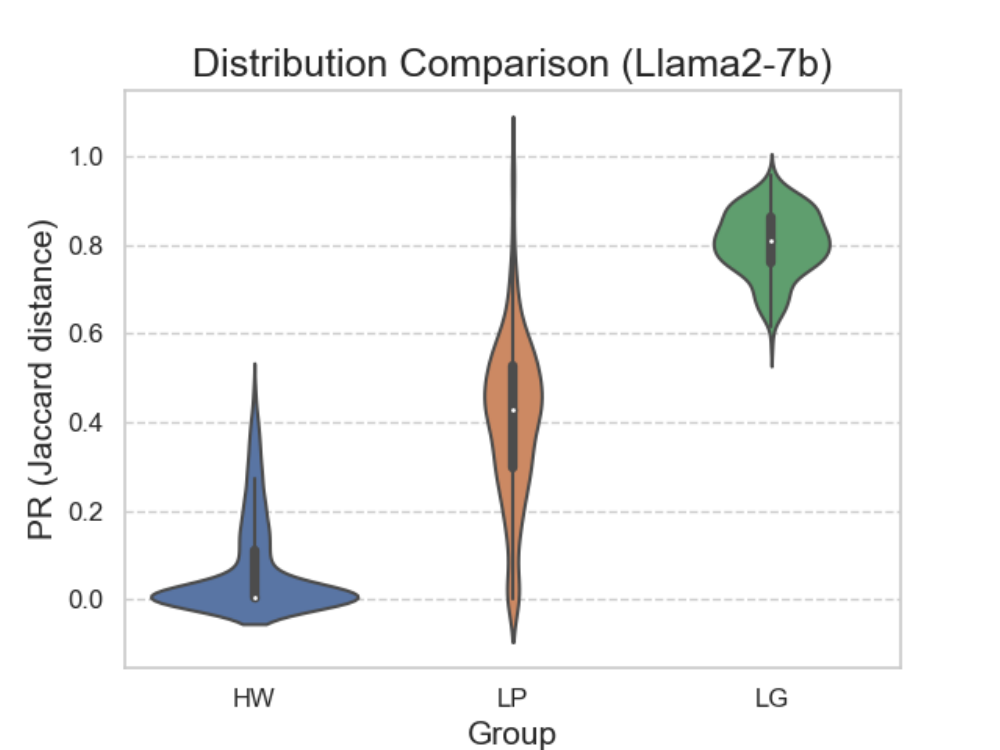}
    \caption{Jaccard distance}
    \end{subfigure}
    \begin{subfigure}[t]{0.48\textwidth}
    \includegraphics[width=\textwidth]{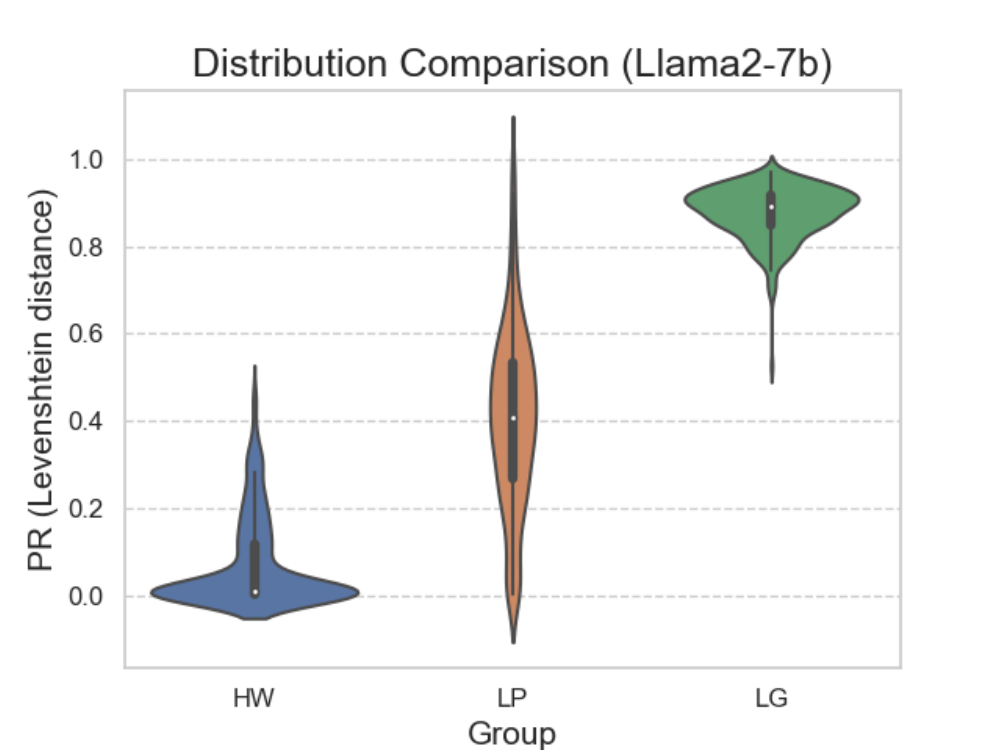}
    \caption{Levenshtein distance}
    \end{subfigure}
    \caption{Differences between predicted PR for human-written texts (HW), Llama-polished texts (LP) and Llama-generated texts (LG).}
    \label{comparison_llama}
\end{figure}

Our proposed "Polish Ratio" method can also be generalizable to other LLMs like Llama2 \cite{touvron2023llama}. 
Following the framework shown in Figure \ref{process}, we replaced the ChatGPT with the Llama2-7b model and polished the same abstracts in HPPT. Under the same experiment settings as reported in section \ref{Reproduction details}, our method can effectively distinguish human-written (HW), Llama-polished (LP), and Llama-generated (LG) texts, as shown in Figure \ref{comparison_llama}. Both results suggest that our proposed method can be generalized to different languages and work on different kinds of powerful LLMs.

\section{Conclusion}
In this paper, we focus on building a more robust ChatGPT detector with independent explanations by detecting the ChatGPT-polished text. We first develop a dataset called Human-ChatGPT Polished Paired abstracTs (HPPT). Then we train a Roberta-based model that can distinguish human-written texts and ChatGPT-polished texts. We also propose a Polish Ratio method to indicate the degree of ChatGPT involvement in the text. Experiments demonstrate that our model achieves better robustness than the baselines. Equipped with the help of the robust detector and Polish Ratio explanation we proposed, users can make an accurate and convincing judgment of the suspected texts. In the future, we will focus on investigating the specific patterns ChatGPT prefers to use from the perspective of some dynamic features of the text.

\section*{Acknowledgement}
This research is supported by the internal project of Shenzhen Science and Technology Research Fund (Fundamental Research Key Project Grant No. JCYJ20220818103001002), and the Internal Project Fund from Shenzhen Research Institute of Big Data under Grant No. T00120-220002.

\bibliographystyle{unsrt}  
\bibliography{custom}  
\end{document}